\documentclass[aps,twocolumn,pra,superscriptaddress,longbibliography,floatfix]{revtex4-2}
\usepackage{amsmath}
\usepackage[T1]{fontenc}
\usepackage{graphicx}
\usepackage{amssymb}
\usepackage{amsfonts}
\usepackage{color}
\usepackage{xfrac}
\usepackage{mathtools}
\def\beq{\begin{equation}}
\def\eeq{\end{equation}}
\begin{document}
\title{Training a neural network with exciton-polariton optical nonlinearity}
\author{A. Opala}
\affiliation{Institute of Physics, Polish Academy of Sciences, Al. Lotnik\'ow 32/46,PL-02-668 Warsaw, Poland}
\email{opala@ifpan.edu.pl}
\author{R.~Panico}
\affiliation{Dipartimento di Matematica e Fisica E. De Giorgi, Universit\`a del Salento, Campus Ecotekne, via Monteroni, Lecce 73100, Italy}
\affiliation{CNR NANOTEC-Institute of Nanotechnology, Via Monteroni, 73100 Lecce, Italy}
\author{V.~Ardizzone}
\affiliation{CNR NANOTEC-Institute of Nanotechnology, Via Monteroni, 73100 Lecce, Italy}
\author{B.~Pi\k{e}tka}
\affiliation{Institute of Experimental Physics, Faculty of Physics, University of Warsaw, ul. Pasteura 5, PL-02-093 Warsaw, Poland}
\author{J.~Szczytko}
\affiliation{Institute of Experimental Physics, Faculty of Physics, University of Warsaw, ul. Pasteura 5, PL-02-093 Warsaw, Poland}
\author{D.~Sanvitto}
\affiliation{CNR NANOTEC-Institute of Nanotechnology, Via Monteroni, 73100 Lecce, Italy}
\affiliation{INFN, Sez. Lecce, 73100 Lecce, Italy}
\author{M.~Matuszewski}
\affiliation{Institute of Physics, Polish Academy of Sciences, Al. Lotnik\'ow 32/46,PL-02-668 Warsaw, Poland}
\author{D.~Ballarini}
\affiliation{CNR NANOTEC-Institute of Nanotechnology, Via Monteroni, 73100 Lecce, Italy}

\begin{abstract}
In contrast to software simulations of neural networks, hardware implementations have often limited or no tunability. While such networks promise great
improvements in terms of speed and energy efficiency, their performance is limited by the difficulty to apply efficient training. We propose and realize experimentally an optical system where highly efficient backpropagation training can be applied through an array of highly nonlinear, non-tunable nodes.  The system includes exciton-polariton nodes realizing nonlinear activation functions. We demonstrate a high classification accuracy in the MNIST handwritten digit benchmark in a single hidden layer system.
\end{abstract}

\maketitle

\section{Introduction}
Over the course of the last decade, artificial intelligence and artificial neural networks have become extremely valuable tools in the industry, research, and everyday life. Deep neural networks excel in many tasks, including image and speech recognition, language processing, or autonomous driving~\cite{LeCun_DeepLearning}. Substantial part of the success can be attributed to the development of effective training algorithms, and the widely applied backpropagation method~\cite{Rumelhart_Backpropagation} in particular. While the effectiveness of existing solutions is unquestionable, it is commonly believed that further progress, in particular in edge computing applications, can only be sustained if software simulations of neural networks are replaced by systems where neural structure of the network is implemented in hardware~\cite{Huang_ReviewNeuromorphic,Misra_ANNSurvey,Grollier_review}. This is dictated by the necessity to develop systems characterized by high speed and high energy efficiency, which is difficult to achieve in the von Neumann computer architecture, where huge amounts of data is transmitted back and forth between memory and computing units.

In recent years, hardware neural networks have been realized in many systems, including CMOS electronics, memristors, and photonic systems~\cite{Merolla,Loihi,furber2014spinnaker,benjamin2014neurogrid,prucnal2017neuromorphic,Wetzstein_review,Grollier_review,Shastri_review,Feldmann_AllOpticalSpikingNetwork,tait2017neuromorphic,Vandoorne,Huang_ReviewNeuromorphic,Lin,antonik2019large,Soljacic_DeepLearning}. In particular, recent realizations using exciton-polaritons in optical microcavities achieved state-of-the-art accuracy in the MNIST handwritten digit recognition benchmark \cite{Opala_NeuromorphicComputing,Ballarini_Neuromorphic,Mirek_Neuromorphic}. Exciton-polaritons are composite quasiparticles that result from strong quantum coupling of semiconductor excitons and cavity photons~\cite{Kavokin_Book,Carusotto_QuantumFluids}. They are characterized by efficient transport via the photonic component and strong interparticle interactions due to the matter component. These properties make them promising candidates for future applications in efficient nonlinear information processing~\cite{Opala_NeuromorphicComputing,Baumberg_SubfemtojouleSwitches,Bramati_SpinSwitches,Sanvitto_TwoFluid,Sanvitto_Transistor,Savvidis_TransistorSwitch,Lagoudakis_RTOrganicTransistor,Baranikov_AllOptical,Liew_Neurons,Espinosa_perceptrons}. Exciton-polariton systems can potentially provide orders of magnitude improvements in speed and energy efficiency as compared to electronics and other optical systems~\cite{Matuszewski_EnergyEfficient}.

\begin{figure*}[ht!]
\centering
  \includegraphics[width=0.7\textwidth,height=13cm]{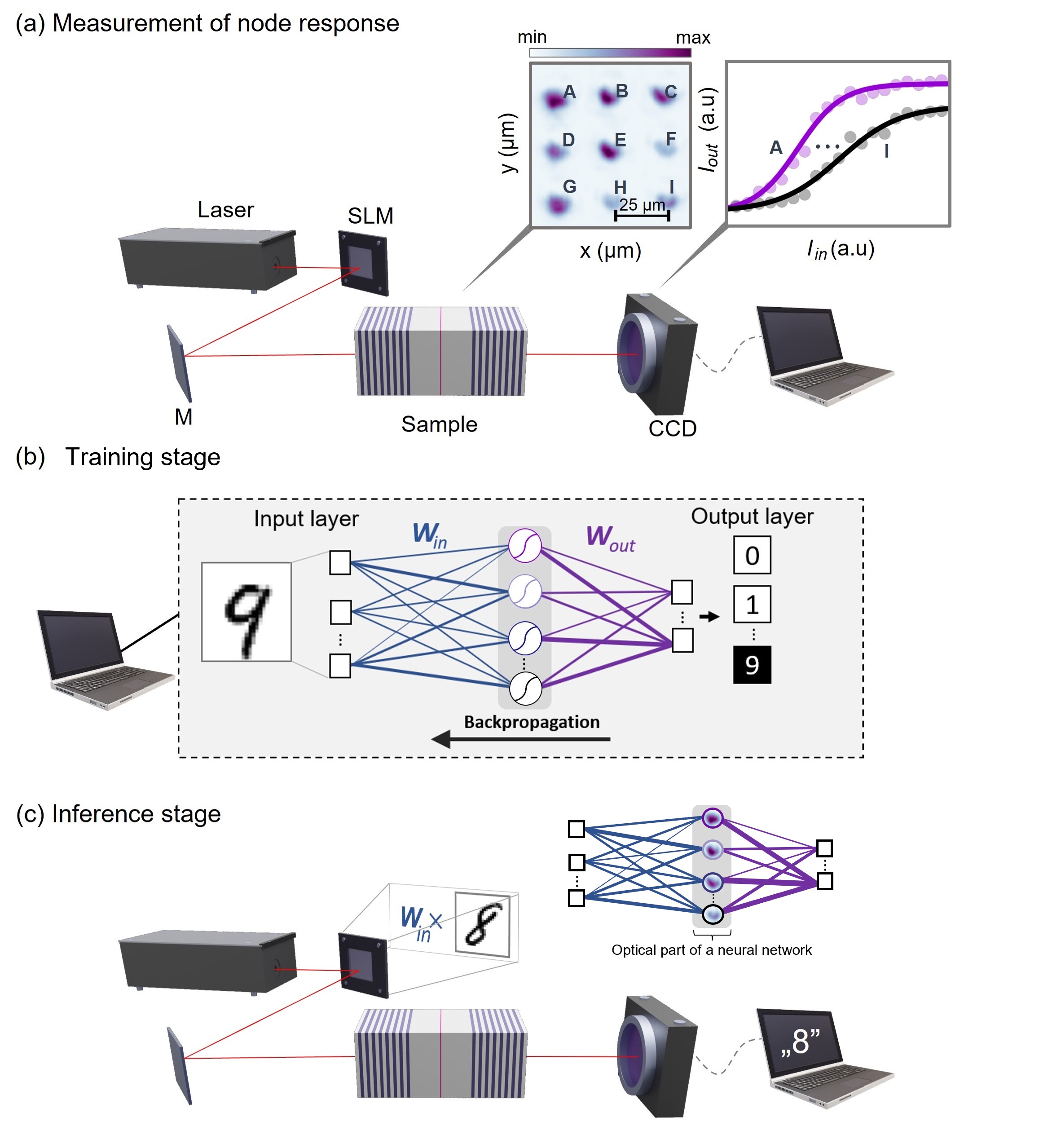}
  \caption{Scheme of the experiment. (a) In the initial stage, we measure the input-output dependence of light transmission through the microcavity. The spatial pattern of input light creates a 3$\times$3 array of uncoupled exciton-polariton nodes characterized by a sigmoid-like response. (b) Training stage is performed in software, where the previously measured node response functions are reproduced to perform backpropagation in a single hidden layer feedforward network. Training results in optimal input weights $W_{\rm in}$ and output weights $W_{\rm out}$. (c) In the inference stage, the obtained weights are used to classify handwritten digits using the nonlinear response of polariton nodes.}
\label{fig:fig_1}
\end{figure*}

A precise, differentiable model of a physical system and its tunability are requisites for the implementation of efficient training algorithms such as backpropagation. In the absence of such a model, one often designs the system according to the reservoir computing  paradigm~\cite{Jaeger_HarnessingESN,Maass_RealTimeComputing,Lukosevicius_RCapproachestoRNN}, where the majority of synaptic weights are unchanged during the training phase~\cite{Vandoorne,Brunner_ParallelPhotonicIPGigabyte,Torrejon,Du_MemristorRC,Tanaka,Larger_HiSpeedReservoirComputing}. On the other hand, a large number of reservoir nodes is usually required to achieve high levels accuracy, and even in the case of very large networks it cannot match the accuracy of backpropagation based networks. 

In recent years, backpropagation has been implemented in several optical systems~\cite{Zhou_LargeScale,McMahon_PAT,Soljacic_DeepLearning,McMahon_SinglePhoton}, however experiments were mostly limited to linear systems where light propagated in free space or in a linear medium. In fact, it is well known that propagation of light is well suited for linear operations such as vector-matrix multiplication~\cite{goodman1978fully,Chang_CNN,Spall_OVMM}.  On the other hand, nonlinearity is a key component of the majority of efficient neural networks. The implementation of nonlinear physical nodes requires a system characterized by a suitable nonlinear response, a precise and differentiable model, reproducibility, stability, and a low amount of noise. All these requirements are difficult to fulfill and their absence strongly influences the accuracy of the predictions of the network. Recently, it was demonstrated that a physics-aware training can lead to improved accuracy~\cite{McMahon_PAT}, but the optimal physical platform is still not known. 

Here, we use the backpropagation algorithm to teach optical neural networks in which nonlinear hardware nodes are non-tunable. We demonstrate a system that includes exciton-polariton nodes exhibiting a strong nonlinear input-output dependence that can be measured precisely. We show that such a precise characterisation of the nodes can be used to perform efficient training. We physically separate the tunable linear weights from the non-tunable nonlinear nodes, whose task is to apply a nonlinear activation function. The uncontrollable and static activation functions of each of the nodes is determined experimentally before the training phase, which allows the application of the backpropagation algorithm offline.

In our proof-of-principle demonstration, we realize a single hidden layer feedforward neural network using optically excited exciton-polariton nodes, where both input and output weights are applied electronically. Despite the experimental imperfections, we achieve the MNIST inference accuracy of 96\%, close to that of a software simulation carried out in the Tensorflow package~\cite{TensorFlow}. This result is similar to the one obtained with physics-aware training method using second harmonic optical nonlinearity in a five-layer network including more than a thousand physical nodes~\cite{McMahon_PAT}. Our polariton network consists of only a few tens of nodes and training does not require hardware gradient estimation, which drastically reduces training time. We emphasize that linear weighting in our network could be realized in principle all-optically, as has been demonstrated previously~\cite{Brunner_ReinforcementLearning, Zuo_AllOpticalNN, chang2018hybrid, farhat1985optical, goodman1978fully,gruber2000planar, lu1989two,Spall_OVMM,Zhou_LargeScale}. Our work opens the way to more complex realizations, including deep and recurrent networks in systems with limited hardware tunability~\cite{Huang_ReviewNeuromorphic}.

\section{Results}

\begin{figure}[ht!]
 \includegraphics[width=0.85\linewidth]{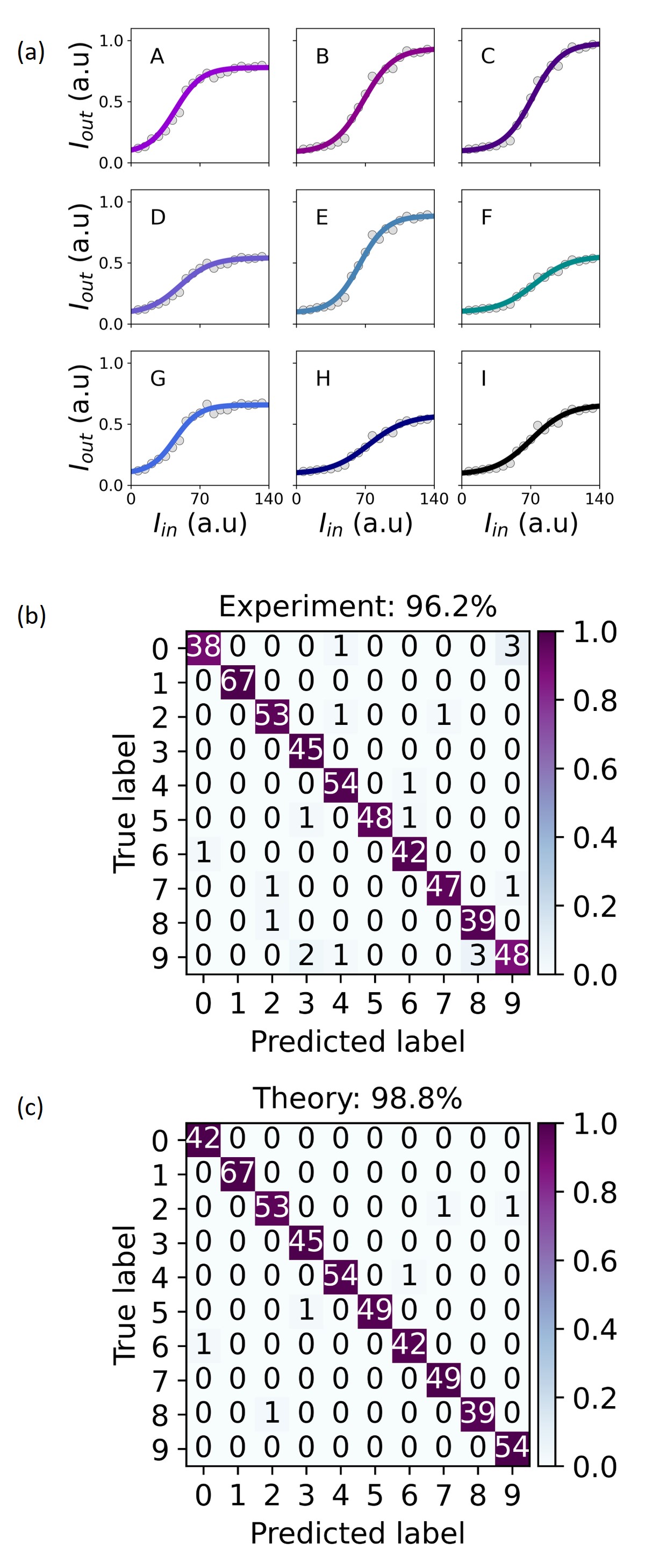}
  \caption{(a) The  measured response of polariton nodes (circles) together with the corresponding analytical fits (solid lines). (b) Confusion matrix resulting from the experimental inference in the case of the testing set. Numbers and color scale indicate the number of samples for a given pair of true and predicted labels. The diagonal corresponds to correct predictions. (c) Confusion matrix for an identical network and identical dataset simulated in the Tensorflow package.}
\label{fig:fig_2}
\end{figure}

Implementation of machine learning typically consists of two separate stages. In the training stage, the system is taught to classify or predict using data from the training dataset. In this stage the synaptic connection weights are tuned, with the aim to increase the accuracy of predictions. The second stage, called the inference (or testing) stage, begins after all training samples have been processed. In this stage, synaptic weights are no longer tuned, and the system is not learning any more. The system is processing a testing dataset that consists of samples that it has not seen before. The accuracy of predictions in the inference stage is the most important benchmark of the network.

While training is usually a time consuming and demanding process, once taught, the system is able to make predictions for an arbitrary number of samples in the inference stage. In many practical applications, it is the inference stage that requires larger amount of resources, and may take indefinitely long time. For example, language processing models can be used to process arbitrarily large number of sentences without any need for readjustment after initial training. Therefore, from the practical point of view, it is valid to seek methods that improve the efficiency of inference, even when not increasing the efficiency of training at the same time~\cite{SurveyAccelerators}.

In our approach we focused on the efficient hardware implementation of  inference. To this end, we add an additional initial stage before training -- the measurement of the response of each physical node -- that allows to construct an accurate software model describing each of the hardware neurons. This allows to perform training in software, while inference is realized in hardware. 
The entire process is  schematically depicted in Fig.~\ref{fig:fig_1}. A standard artificial neuron realizes two functions: (i) tunable synaptic weighting of inputs and (ii) a non-tunable activation function.  We separate physically function (i) from  function (ii), the latter being realized by a set of non-tunable nonlinear nodes. In our case, these nodes are optically excited exciton-polariton modes of an optical microcavity. The knowledge of node response, measured in the initial stage as depicted in Fig.~\ref{fig:fig_1}(a), allows to perform the training stage entirely in software, using the backpropagation method as shown in Fig.~\ref{fig:fig_1}(b). The resulting synaptic weights are implemented physically only in the inference stage, as in Fig.~\ref{fig:fig_1}(c). We choose a simple single-hidden-layer feedforward neural network model to facilitate the backpropagation training procedure. Importantly, in contrast to reservoir computing method, we adjust {\it all} synaptic weights of the network, including those in the input layer. We will demonstrate that this allows to significantly improve the accuracy of predictions. 

Figure~\ref{fig:fig_1}(a) presents the scheme of the experimental setup. A phase-only spatial light modulator (SLM) is used to modulate the intensity profile of a laser beam into an array of bright spots with individually tunable intensity. These spots, distributed in a 3$\times$3 square lattice, are imaged on the surface of a semiconductor microcavity at resonance with the exciton-polariton energy. Due to the significant exciton-polariton nonlinearities, the light intensity transmitted through the microcavity follows a sigmoid-like behavior as a function of the input intensity~\cite{Ballarini_Neuromorphic}, as shown in the inset of Fig.~\ref{fig:fig_1}(a). The light intensity of each polariton node, labeled (A,B,C,..I) in the figures, is measured by a CCD camera and collected on a computer. Each node is spatially separated from the others to ensure that the output intensity of a polariton node depends only on its input intensity, independently from the input intensity of the other polariton nodes. This configuration is chosen for simplicity, as it facilitates the training procedure.
However, our method is not limited to the case of isolated nodes. In the case of interconnected nodes, the system could be more accurate, even if requiring longer times and more complex methods to perform training.

The training stage, depicted in Fig.~\ref{fig:fig_1}(b), is realized entirely in software. We use the Tensorflow package to simulate a single-layer feedforward neural network shown schematically in the Figure. We consider the MNIST handwritten digit dataset, which contains 60 000 samples in the training set and 10 000 samples in the testing set, each sample being a 28$\times$28 grayscale image of a digit between 0 and 9 and the corresponding label~\cite{LeCun_MNIST}. The number of nodes in the hidden layer of the network is equal to $9M$, where $M$ is the number of experimental shots that will be used to process a single digit in the inference stage. The activation function of each hidden node is given by the node response  measured in the initial stage. In the output layer, the softmax function is used to choose the class corresponding to one of the ten digits.  We use the backpropagation algorithm to teach the network, which provides optimal values of synaptic weights in both in the hidden layer $W_{\rm in}$ and output layer $W_{\rm out}$ connections.

In the inference stage, shown in Fig.~\ref{fig:fig_1}(c), we use the spatial light modulator to encode input data from the testing set. Additionally, data is multiplied by input synaptic weights, which is in our case realized in software. Light intensity incident on each microcavity node is a sum of inputs multiplied by the corresponding input synaptic weights. The role of the microcavity is to apply the nonlinear activation function via exciton-polariton interactions. The transmitted light corresponds therefore to the neuron outputs. Further, the intensity measured on a CCD camera is collected, multiplied by output synaptic weights in software and a softmax function is used to determine the predicted class. 

\begin{figure}[ht!]
\centering
\includegraphics[width=0.85\linewidth]{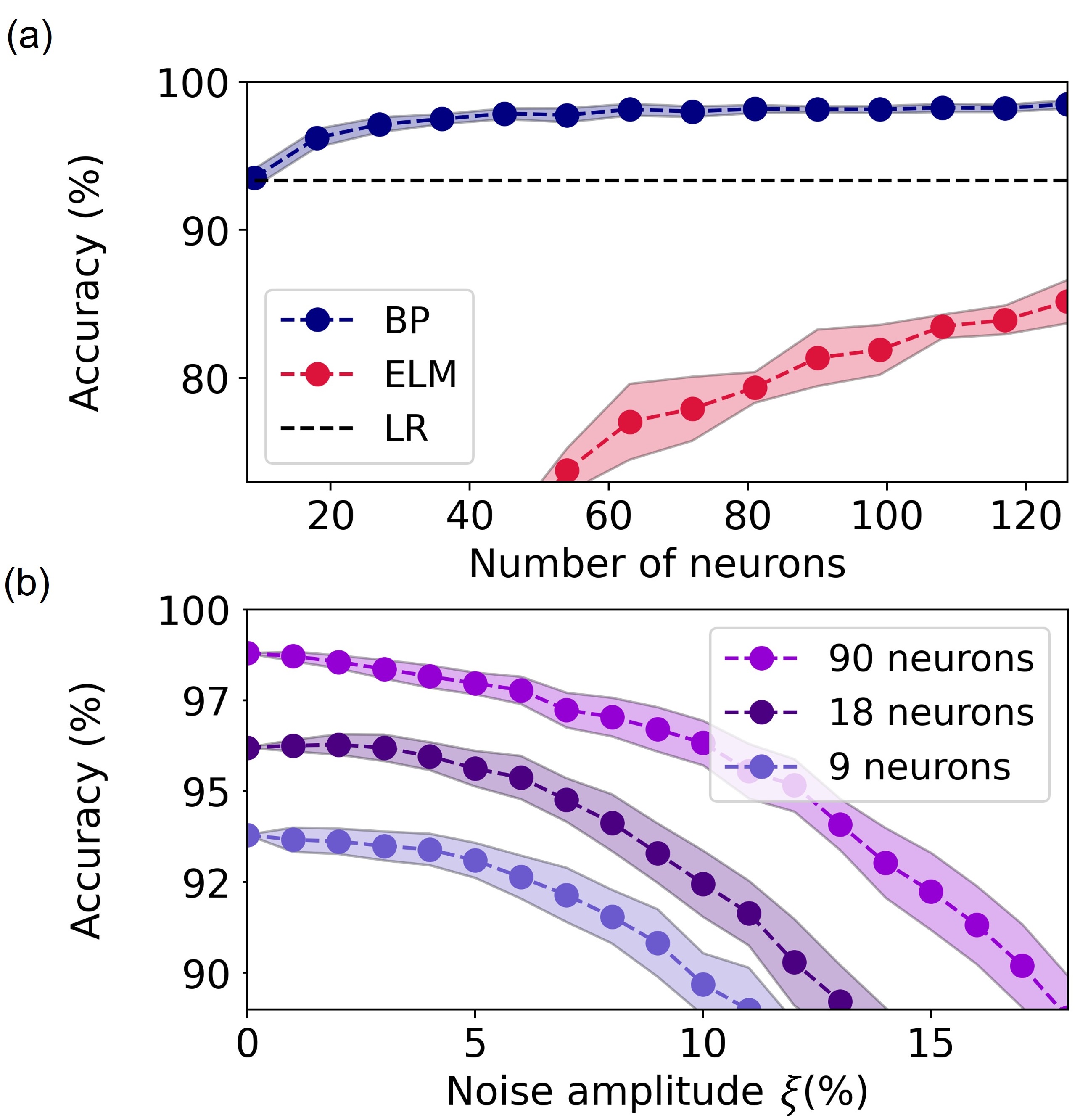}
\caption{(a) Numerically estimated inference accuracy of backpropagation networks (BP), compared to "extreme learning machine" networks (ELM), where only the output weights are adjusted, and the linear classification using logistic regression (LR). Horizontal axis corresponds to the number of neurons in the hidden layer. (b) Accuracy of inference in function of the amplitude of noise.}
\label{fig:fig_3}
\end{figure}

In Figure~\ref{fig:fig_2}(a), we show the measured response of each polariton node, together with the corresponding analytical fits. The analytical functions are necessary to apply the backpropagation training method, which requires the knowledge of the derivative of an activation function in order to calculate the update of input synaptic weights $w_{ij}^{in}$ according to the formula

\begin{equation} \label{Eq:1}
    \Delta w_{ij}^{\rm in}=-\eta\sum_{k=1}^{M}(y_k-d_k)\frac{dy_k}{d v_i}\frac{dv_i}{d w_{ij}^{\rm in}}x_j,
\end{equation}
where $\eta$ is the learning rate, ${\bf{x}}=[x_1,x_2,x_3,\dots,x_N]^T$ and ${\bf{y}}=[y_1,y_2,y_3,\dots,y_M]^T$ are the input and output vectors, respectively, and ${\bf{d}}=[d_1,d_2,d_3,\dots,d_M]^T$ is the target output vector. Activations of neurons in the hidden layer are given by the equation $v_i=\varphi_i(\sum_{j=0}^Nw_{ij}^{\rm in} x_j)$, where $\varphi_i(I^{\rm in}_i)$ is an activation function and $N=28^2$ is the number of inputs. We approximate the polariton response by sigmoid analytical activation functions 
\begin{equation} \label{Eq:2}
    \varphi_i=\frac{a_i}{1+\exp(-(I^{\rm in}_i-t_i)/c_i)}+d_i,
\end{equation}
where $a_i$, $t_i$, $c_i$ and $d_i$ are parameters obtained by fitting to the experimental data, where $i=\{1,2,3, \cdots, 9M\}$. Overall, we fit 9 sigmoid functions to the response of 9 polariton nodes in the $(3\times3)$ network. Figure~\ref{fig:fig_2}(b) shows the results of the experimentally obtained accuracy at the inference stage, i.e.~predictions for 500 digits from the testing set. For comparison, in  Figure~\ref{fig:fig_2}(c) we show the accuracy obtained in a Tensorflow software simulation of the same network and with the same dataset. The accuracy achieved in the experiment of 96.2\% is comparable to the result obtained with a binarized network in~\cite{Mirek_Neuromorphic} and higher than that obtained in a reservoir computing approach with a similar number of nodes~\cite{Ballarini_Neuromorphic}. In comparison to the binarized network~\cite{Mirek_Neuromorphic}, the setup presented here is much simpler, the number of nodes is much lower (tens instead of tens of thousands), and there is no need to create optical binary gates first. Consequently it has the potential to achieve higher speed of data processing and is more scalable. 

We emphasize that this excellent result has been obtained with a relatively small network, including only 90 nodes in the hidden layer ($M=10$). This results from the use of backpropagation, which adjusts synaptic weights in all the network layers. Achieveing a similar accuracy with a reservoir computing network requires a much larger number of nodes in the hidden layer~\cite{Opala_NeuromorphicComputing}.
To investigate the advantage of our approach in more detail, in Fig.~\ref{fig:fig_3}(a) we compare the accuracy of our network, simulated in software, to the accuracy of an "extreme leraning machine" (ELM) network~\cite{Huang_ExtremeLearningMachines,Conti_ELM}. The latter network has an identical architecture to ours, however the input synaptic weights $W_{\rm in}$ are random and not adjusted in the training phase. Since only the synaptic weights in the output layer are adjusted, it can be considered a simplified feedforward analog of a reservoir computing network. Additionally, we compare these results with the accuracy level obtained using a linear classification method (taught using logistic regression), where nonlinear transformation of data is absent. It is clear that the use of backpropagation to adjust the weights results in a significant improvement of accuracy as compared to the ELM network, and in contrast to reservoir computing, it surpasses the linear classification method even for a network with a very small number of nodes.

We also consider the effect of noise on the performance of the network. In Figure~\ref{fig:fig_3}(b) we show accuracy as a function of the amplitude of noise. We add to the activation functions $\varphi_i$  random variables $\delta \varphi_i$, $\tilde{\varphi_i} = \varphi_i + \delta \varphi_i$, that have Gaussian distributions with zero mean and standard deviations equal to 
\begin{equation} \label{Eq:3}
\sigma(\delta \varphi_i)=\xi \Delta_i,    
\end{equation}
where $\Delta_i=\max(I_i^{\rm out})-\min(I_i^{\rm out})$ is the amplitude of the i-th polariton neuron response and $\xi$ is the relative amplitude of noise. The accuracy typically remains large if the noise is weak, even in the case of small neural networks. Only large amplitude noise can significantly deteriorate the network performance, while larger networks appear to be more resistant to strong noise.

\section{Discussion}

Exciton-polaritons allow to achieve high speed and energy efficiency of all-optical computation~\cite{Matuszewski_EnergyEfficient}. Here, light intensity per neuron measured before the sample was approximately 1 mW (or 9 $\mu$W/$\mu$m$^2$ with 6 $\mu$m node radius). For comparison, the intensity of the control beam in Zuo et al.~\cite{Zuo_AllOpticalNN} was up to 50 $\mu$W per node. However, the most important figure of merit is the energy cost, which depends also on the data rate of the device. In polariton systems, the natural time scale is of the order of picoseconds, which makes it possible to achieve optical nonlinearities at the cost of femtojoules. For comparison, the energy cost of nonlinear activation based on phase change materials was a few hundred picojoules~\cite{Feldmann_AllOpticalSpikingNetwork}. One could also compare the energy cost to electronics, where the highest energy cost arises from linear operations, and is of the order of 100 pJ per MAC in GPU systems and 10 pJ per MAC in TPU systems in the case of fully connected networks~\cite{Jouppi_InDatacenter}. On the other hand, the above estimates for optical systems do not take into account optical signal generation, detection, and additional electronic elements of the system. The energy cost of these needs to be reduced to achieve a high overall system efficiency.

We emphasize that while in our experimental implementation the tunable weights are implemented in software, there are known all-optical methods for vector-matrix multiplication. These methods have been realized in many experiments, both in the case of coherent and incoherent light~\cite{Brunner_ReinforcementLearning, Zuo_AllOpticalNN, chang2018hybrid, farhat1985optical, goodman1978fully,gruber2000planar, lu1989two}. In particular, multiplication of a vector by a matrix containing 3 000 elements was implemented recently~\cite{Spall_OVMM}. Thus, the inference stage could be realized in an all-optical system, without use of any electronic elements. Such an optical network could be completely passive, as it would not require any external power supply except for the laser source. Room temperature polaritons can be used to avoid the cost of sample cooling~\cite{Grandjean_RTPolaritonLasing,Fieramosca_perovskites,Malpuech_ZnOCondensate}. Moreover, replacing the spatial light modulator with ultrafast modulators working at the GHz data rate, one could take advantage of the very short, picosecond  timescales of the optical system. Another possibility is to use an on-chip integrated version of the system, where transmission is tuned by optoelectronic modulators using the Stark effect acting on the exciton component~\cite{Sanvitto_Stark}. Physics-aware backpropagation training~\cite{McMahon_PAT} could be also used to perform backpropagation, possibly leading to considerable improvement of the accuracy in a multilayer system.

\acknowledgments
MM acknowledges support from National Science Center, Poland grant 2017/25/Z/ST3/03032 under the QuantERA program. AO acknowledges support from National Science Center, Poland grant 2020/36/T/ST3/00417. BP acknowledges support from National Science Center, Poland grant 2020/37/B/ST3/01657.

\appendix

\section{Model}

Our implementation is a dense feed-forward neural network that contains a single hidden layer between the input and output layers. The hidden layer contains $i=9M$ neurons, which activations were obtained by fitting functions (\ref{Eq:2}) to the measured polariton node responses. The particular activation function $\varphi_i$ depends on the physical node $(A,B,C,\cdots I)$ that it corresponds to. 
The neural network transforms the input vector $\bf{x}$ into the output $y(\bf{x})$ according to the following equation
\begin{equation}
\label{Eq:4}
y({\bf{x}})=f({\bf{W}}_{\rm out}\phi({\bf{W}}_{\rm in}{\bf{x}}+{\bf{b}}_{\rm in})+{\bf{b}}_{\rm out})
\end{equation}
where $f(\bf{\cdot})$ is the softmax function, ${\bf{W}}_{\rm out}$  is the weight matrix of connections between the output and the hidden layer. Function $\phi$ applies neuron activation functions  $\phi({\bf{I^{\rm in}}})=[\varphi_1(I_1^{\rm in}), \varphi_2(I_2^{\rm in}), \varphi_3(I_3^{\rm in}), \dots, \varphi_{9M}(I_{9M}^{\rm in})]$. Matrix ${\bf{W}}_{\rm in}$ contains the weight between the input and the hidden layer, while ${\bf{b}}_{\rm in}$ and ${\bf{b}}_{\rm out}$ are the input and output bias vectors, respectively.
We optimise the weight matrices and bias vectors using the ADAM optimiser. We use the supervised learning method and the backpropagation algorithm. The application of the backpropagation method is possible thanks to the simulation of neuron activations by sigmoid functions, which are continuous and differentiable.

\section{Methods}

The semiconductor microcavity used in these experiments is a planar cavity with three 8 nm In$_{0.04}$Ga$_{0.96}$As quantum wells embedded between two AlAs/GaAs distributed Bragg reflectors, and kept at a temperature of $\sim7$ K. The high quality-factor ($Q \sim 10^4$) of this sample results in a polariton lifetime of 10 ps. In particular the region of the sample we used exhibits an exciton-cavity detuning of $\delta\simeq1.7$ meV and polaritons are pumped by a continuous wave laser tuned at $\lambda=836.64$ nm.

To shape the profile of the laser beam we used a spatial light modulator (SLM), a liquid crystal display with a surface area of approximately 2 cm$^2$ and a 1920$\times$1080 resolution. Applying a voltage to the cells changes the orientation of the liquid crystals and in turn the effective refractive index seen by the incident light. The control of the birefringence of each pixel allows us to spatially design the amplitude and phase of the reflected wave.

To create the node lattice on the sample we reconstructed the real space image of the SLM, reduced in size by a factor of 50. On top of the displayed lattice pattern we used a blazed grating pattern that serves two purposes: on the one hand it allows us to block the zeroth-order reflection from the SLM that brings the non-modulated part of the laser, on the other hand by changing the diffraction efficiency of the grating we are able to tune the intensity of the individual nodes. Furthermore a phase difference of $\pi$ is applied between each node, for a better separation.

The pattern is focused onto the microcavity sample by a camera objective
with a focal length of 5 cm. The pump beam frequency is slightly
blue-detuned with respect to the polariton resonance: increasing the
pump power results in an energy shift of the polariton resonance due to
the exciton-mediated interactions. The resulting transmission intensity
is a highly nonlinear function of the pump power, realizing a physical
implementation of a sigmoidal response function.
The emission is collected with a 2 cm aspheric lens and recorded on a
coupled charge device (CCD).

\bibliographystyle{apsrev4-2}
\bibliography{bibliography}

\end{document}